\documentclass[conference]{IEEEtran}
\IEEEoverridecommandlockouts
\usepackage{cite}
\usepackage{amsmath,amssymb,amsfonts}
\usepackage{algorithmic}
\usepackage{graphicx}
\usepackage{textcomp}
\usepackage{xcolor}
\def\BibTeX{{\rm B\kern-.05em{\sc i\kern-.025em b}\kern-.08em
    T\kern-.1667em\lower.7ex\hbox{E}\kern-.125emX}}

\begin{document}

\title{Numerical Exploration of Training Loss Level-Sets \\ in Deep Neural Networks
\thanks{This work was supported in part by DARPA award number HR00111890044.}
}


\author{
\IEEEauthorblockN{
Naveed Tahir}
\IEEEauthorblockA{\textit{Elec. Engr. \& Comp. Sci.}}
\textit{Syracuse University}\\
Syracuse, NY, USA \\
\texttt{ntahir@syr.edu}
\and
\IEEEauthorblockN{
Garrett E. Katz}
\IEEEauthorblockA{\textit{Elec. Engr. \& Comp. Sci.}}
\textit{Syracuse University}\\
Syracuse, NY, USA \\
\texttt{gkatz01@syr.edu}
}

\maketitle

\begin{abstract}
We present a computational method for empirically characterizing the training loss level-sets of deep neural networks. Our method numerically constructs a path in parameter space that is constrained to a set with a fixed near-zero training loss. By measuring regularization functions and test loss at different points within this path, we examine how different points in the parameter space with the same fixed training loss compare in terms of generalization ability. We also compare this method for finding regularized points with the more typical method, that uses objective functions which are weighted sums of training loss and regularization terms. We apply dimensionality reduction to the traversed paths in order to visualize the loss level sets in a well-regularized region of parameter space. Our results provide new information about the loss landscape of deep neural networks, as well as a new strategy for reducing test loss.
\end{abstract}


\section{Introduction}
Recent advances in neural networks have led to impressive but often inexplicably high performance.  For example, networks trained with local optimization often converge to globally optimal (zero) training loss.  Our incomplete theoretical understanding of how neural networks operate (and when they might fail) is often described as an ``explainability'' or ``interpretability'' problem in AI.

Inexplicable AI often translates to low user adoption in high stakes applications such as self-driving cars, medical diagnosis, and criminal justice.  In response, researchers have recently sought to strengthen our mathematical understanding of how and when various neural architectures perform well.  For example, Poole et al. \cite{poole2016exponential} showed that global curvature of a neural network function increases exponentially with depth but not width, to some extent explaining why deep architectures perform so well (although cf. \cite{zagoruyko2016wide}). Soudry et al. \cite{DBLP:journals/corr/SoudryC16} explained performance from a different perspective, proving that when the topmost layers of a network are sufficiently large relative to the training data, then any differentiable local minimum is in fact a global minimum with zero training error.  Conversely, Baldi et al. \cite{baldi2019capacity} showed that the number of boolean functions that can be represented exactly by feed-forward networks is a cubic polynomial of the layer sizes.  Using different techniques than in \cite{DBLP:journals/corr/SoudryC16}, Du et al. \cite{du2019gradient} also show that gradient descent converges to zero error for various network architectures.

Related to these theoretical results, various practical techniques can ``easily'' achieve negligible training error when the hidden layers and weights are set up correctly.  For example, reservoir computing \cite{jaeger2004harnessing}, extreme learning machines \cite{huang2006extreme}, and some applications of the neural engineering framework \cite{eliasmith2004neural}, have all used techniques where weights and/or activity patterns leading up to the penultimate layer are chosen randomly, and then ``decoding'' or ``read-out'' by the output weights is achieved by solving a linear system with near-zero residual.

Given that real-world neural networks are mostly smooth functions, and often over-parameterized relative to the training data, we can expect that the optima are not isolated points in parameter space, but rather, comprise smooth manifolds (perhaps modulo some singular points). In this paper we seek to empirically characterize such manifolds to improve our understanding and engineering of neural networks.  For example, are the level-sets for a fixed training loss typically path connected? What is the distribution of various regularization values and test loss within the level-set?  For a given regularizer, how does the most regularized point in the level-set compare to those found in the standard way, by optimizing a weighted sum of training loss and regularization term?

To begin answering such questions, we use numerical methods to construct randomly sampled paths within manifolds for fixed near-zero levels of training loss. We focus on near-zero, rather than exactly zero, training loss for two reasons: (1) in practice numerical optimization will not reach true zero loss, and (2) the gradient is zero at true zero loss, which would require higher than first-order methods for traversal. Our approach can be described in terms of two phases.  First, we use standard optimization from random initial weights to minimize the training loss, thereby locating a starting point in the manifold corresponding to a near-zero fixed training loss.  Second, we use mathematical principles from gradient projection and numerical path following \cite{rosen1960gradient,allgower1997numerical} to enumerate a sequence of points within the manifold along a path of interest (for example, one that minimizes a regularization function).  These two phases are repeated many times with random initial conditions to form a random sample of paths within level sets of fixed training loss.  In addition, we collect an empirical distribution of test loss and regularization scores, among other metrics, along these paths to better characterize these level sets.  This experimental setup is used on an illustrative toy example as well as multiple real-world datasets including MNIST \cite{lecun1998gradient}, CIFAR10 \cite{Krizhevsky09learningmultiple}, Iris \cite{fisher1936use}, and Auto-MPG \cite{quinlan1993combining} datasets. We also compare our method against the more typical approach where the objective is a weighted sum of training loss and a regularization term. The results show that our method effectively locates regularized points within the training loss level-sets that better generalize to test data.

\section{Related Work}
The connectedness, or lack thereof, of the training loss landscape has been an interesting theme of exploration in recent studies on deep neural networks. Large (deep and wide) networks embody a certain redundancy which has been argued to cause lowered energy (i.e., loss) barriers and increased connectivity in the loss landscape \cite{sagun2018empirical, venturi2019spurious, DBLP:conf/icml/Nguyen19, DBLP:conf/icml/DraxlerVSH18}. One consequence of such redundancy, for instance, is that slight perturbations to individual parameters in large networks can be offset by making the other parameters adapt in such a way that the any change in the training loss is negligible. Such insights forms the basis of various pruning methods \cite{liu2017learning, lecun1990optimal} as well as the more popular Dropout method \cite{srivastava2014dropout}.

More generally, a similar adapting procedure can be used to find new, and perhaps neighboring, points in the loss landscape with comparable performance. Recently, Garipov et al. \cite{garipov2018loss} and Draxler et al. \cite{DBLP:conf/icml/DraxlerVSH18} independently showed that for multi-layer neural networks it is possible to path-connect a pair of distinct minima in the parameter space. The former show that a pair of distinct sets of weights, obtained, for example, by training two randomly initialized instances of the same network, can be connected through polygonal curves along which the training loss remains nearly the same as that on the distinct end-points. The latter show that the same connection can also be established using the Nudged Elastic Band method \cite{jonsson1998nudged}, a method used to discover reaction pathways in chemical reactions. Similarly, Nguyen \cite{DBLP:conf/icml/Nguyen19} showed that every sublevel-set of an over-parametrized feed-forward network is connected, under mild assumptions on the topology of the network. Our work complements and extends these works in several ways. First, we propose a numerical, architecture-agnostic method for traversing loss level-sets with respect to a given direction of interest (such as one that minimizes a regularizer). Such a traversal requires a single starting point and does not require any of the consecutive points on the level-set to be known beforehand (as is the case in \cite{garipov2018loss} and in \cite{DBLP:conf/icml/DraxlerVSH18}). Second, the numerical nature of the method allows it to sidestep some of the assumptions (such as those on the layer width and topology of the network deduced in \cite{DBLP:conf/icml/Nguyen19} and \cite{venturi2019spurious}) as the traversal is carried out. 

\section{Approach}
\label{sec:numapproach}

We consider smooth neural network models with trainable parameters $\theta\in\mathbb{R}^N$, where $N$ is the number of parameters, along with a non-negative training loss function $\mathcal{L}:\mathbb{R}^N\rightarrow\mathbb{R}_{\geq 0}$ and a regularization function $\mathcal{R}:\mathbb{R}^N\rightarrow\mathbb{R}_{\geq 0}$. Here $\theta$ can be interpreted as a vector obtained by concatenating the flattened weight matrices and bias vectors that span multiple layers in a network.  We are interested in characterizing a level set with a fixed near-zero training loss $\epsilon > 0$.  We denote this level set by $\mathcal{M}$, defined as:
\begin{align}
\mathcal{M} = \{\theta\in\mathbb{R}^N : \mathcal{L}(\theta) = \epsilon\}.
\end{align}
In particular, we are interested in how $\mathcal{R}(\theta)$ varies within $\mathcal{M}$. We focus on the case where $\epsilon > 0$ is near, but not exactly, a minimum value of $\mathcal{L}$.  In this case, $\nabla_\theta \mathcal{L} \ne 0$, and $\mathcal{M}$ is a smooth manifold, with implicit equation $\mathcal{L}(\theta)=\epsilon$.  An initial point $\theta_0\in\mathcal{M}$ can be found with standard gradient-based optimization.  Then we can numerically generate a sequence of discrete steps $\theta_0$, $\theta_1$, ..., $\theta_n$, ... in $\mathcal{M}$ as follows.

Expanding around the current point $\theta_n\in\mathcal{M}$, a second-order Taylor approximation of $\mathcal{M}$'s implicit equation is
\begin{align}
\mathcal{L}(\theta_n+\delta) &\approx \mathcal{L}(\theta_n) + \delta^\top\nabla_{\theta}\mathcal{L}(\theta_n) + \delta^\top H(\theta_n)\delta/2\label{eq:taylor}
\end{align}
where $H(\theta)$ is the Hessian of $\mathcal{L}(\theta)$.  We seek a numerical step $\theta_{n+1}=\theta_n+\delta$ that remains in $\mathcal{M}$, i.e., $\mathcal{L}(\theta_{n+1}) = \mathcal{L}(\theta_n) = \epsilon$.  From (\ref{eq:taylor}) it is apparent that such a $\delta$ should satisfy
\begin{align}
0 &\approx \delta^\top\nabla_{\theta}\mathcal{L}(\theta_n) + \delta^\top H(\theta_n)\delta/2\label{eq:taylor0}
\end{align}
We consider the case when $\mathcal{M}$ is a level set of training loss near, but not at, a local minimum.  In this case, the first-order Taylor term dominates, and we seek a $\delta$ satisfying
\begin{align}
\delta^\top \nabla_\theta\mathcal{L}(\theta_n) = 0.\label{eq:Lnull}
\end{align}
An infinitesimal step in direction $\delta$ satisfying the condition above will remain in $\mathcal{M}$. More interestingly, there is some freedom in choosing $\delta$ as long as it satisfies Eq. \ref{eq:Lnull}. Let $r$ denote some direction of interest that may or may not be confined to $\mathcal{M}$.  For example, $r$ could be $-\nabla_\theta\mathcal{R}(\theta_n)$, in which case we seek to minimize the regularization term, or it could be sampled from a multidimensional normal distribution, in which case we will proceed on a random walk.  

Given $\nabla_\theta \mathcal{L}$ and some such $r$, we therefore seek a $\delta$ with a non-zero component along $r$ that also satisfies Eq. \ref{eq:Lnull}, which is a linear problem.  However, a numerical step along $\delta$ is not infinitesimal and will introduce some error (i.e., $\mathcal{L}(\theta_n + \delta) \ne \epsilon$) that must be corrected numerically.  To that end, we adopt a predictor/corrector scheme \cite{allgower1997numerical} for phase 2 of our method.  The predictor step advances along a $\delta$ that is orthogonal to $\nabla_\theta \mathcal{L}$ and has a component along $r$.  If this introduces any non-negligible deviation in $\mathcal{L}$, corrector steps are then used to return to $\mathcal{M}$.  More formally, the loss deviation $\mathcal{D}$ for the $n^{th}$ iteration of phase 2 is defined below:
\begin{align}
\begin{split}
\mathcal{D}(\theta_n) = (\mathcal{L}(\theta_n) - \mathcal{L}(\theta_0))^2
\label{eq:lossdeviation}
\end{split}
\end{align}
where $\theta_0$ is the traversal starting point in $\mathcal{M}$ found by the first phase. In a predictor step, while the training loss deviation is below a pre-specified small threshold, the network weights will be updated according to the following relations:
\begin{align}
\begin{split}
\delta_p &= \nabla_{\theta}\mathcal{R} - \text{proj}_{\nabla_{\theta}\mathcal{L}}\nabla_{\theta}\mathcal{R} \\
\mathcal \theta_{n+1} &= \theta_{n} - \eta\hat{\delta_p} 
 \label{eq:predictor}
\end{split}
\end{align}
where $\hat{\delta_p}$ is the unit vector along the prediction direction ${\delta_p}$, $\eta$ is the learning rate, and $\text{proj}_{b}a$ is the projection of a vector $a$ on to another vector $b$, defined as:
\begin{align}
\text{proj}_{b}a = (a^\top\hat{b})\hat{b}  \label{eq:vectorprojection}
\end{align}
A practical choice of $\mathcal{R}$ is the squared L2 norm: $\mathcal{R}(\theta) = \theta^\top\theta$.

Note that equation \eqref{eq:predictor} differs from standard gradient descent, in that we project away the component of the regularization gradient $\nabla_{\theta}\mathcal{R}(\theta_n)$ along the training loss gradient $\nabla_{\theta}\mathcal{L}(\theta_n)$. Since the resulting $\hat{\delta_p}$ has no components along $\nabla_{\theta}\mathcal{L}(\theta_n)$, taking an infinitesimal step along this direction will not change $\mathcal{L}(\theta)$ but will change (reduce) $\mathcal{R}(\theta)$. However, in practice, numerical steps are not infinitesimal and small errors would accumulate without a corrector step. 

In the corrector step, which sets in if the training loss deviation $\mathcal{D}$ is above the aforementioned threshold, the deviation is minimized using gradient descent. In order to avoid undoing the work in the predictor step, we constrain the corrector steps to move orthogonally to $\delta_p$.  This is done by computing a descent direction -$\nabla \mathcal{D}$ that reduces the loss deviation, and then projecting away its component along $\delta_p$.  The resulting weight update equations for a corrector step, therefore, are:

\begin{align}
\begin{split}
\delta_c &= \nabla_{\theta}\mathcal{D}- \text{proj}_{\delta_p}\nabla_{\theta}\mathcal{D}\\
\mathcal \theta_{n+1} &= \theta_{n} - \eta\hat{\delta_c} 
 \label{eq:corrector}
\end{split}
\end{align}

It remains to identify stopping criteria for this predictor/corrector procedure. Two reasonable stopping criteria include the number of total predictor/corrector steps allowed, and the vectors ${\nabla_{\theta}\mathcal{L}}$ and $\nabla_{\theta}\mathcal{R}$ being anti-parallel. While the total number of steps criterion is a simple resort in case of limited available compute, the anti-parallel criterion is informed by the method of Lagrange multipliers \cite{boyd2004convex} using which the problem of manifold traversal can also be described as a constrained optimization problem. More precisely, for a given $\mathcal{L}$, $\epsilon$ and $\mathcal{R}$, we are interested in traversing $\mathcal{M}$ along a path that minimizes $\mathcal{R}$, i.e.:
\begin{align}
\text{minimize }\mathcal{R}(\theta), \text{ subject to } \mathcal{L}(\theta) = \epsilon. \label{eq:constrainedopt}
\end{align}
Using the method of Lagrange multipliers, we have
\begin{align}
F(\theta, \lambda) &= \mathcal{R}(\theta) + \lambda(\mathcal{L}(\theta) - \epsilon) \label{eq:lagrange}
\end{align}
where $\lambda$ is the multiplier and $F$ is the Lagrangian. The solution to Eq. \ref{eq:lagrange} involves finding stationary points $\theta$ where $\nabla_\theta F = \nabla_\theta\mathcal{R} + \lambda\nabla_\theta\mathcal{L} = 0$. At a point minimizing $\mathcal{R}$, where the constraint is also satisfied, the gradients $\nabla_\theta\mathcal{L}$ and $\nabla_\theta\mathcal{R}$ will be anti-parallel. This is because a decrease in the regularizer $\mathcal{R}$ translates into an increase in $\mathcal{L}$ and vice versa.

\begin{figure}[t]
\centering
\includegraphics[width=0.9\columnwidth]{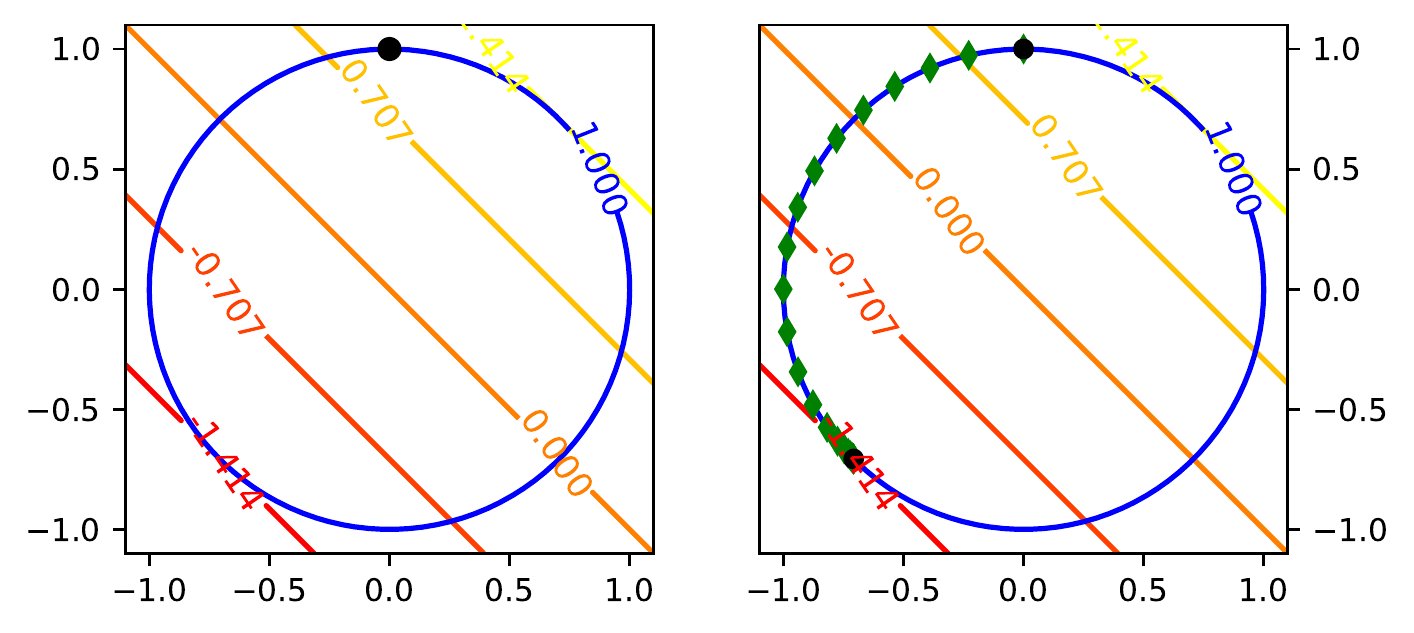}
\caption{\textbf{Left:} The contours for the objective and the constraint given in Eq. \ref{eq:toy_npf}. The black dot at the top of the unit circle is a random starting point for the predictor/corrector procedure. \textbf{Right:} The new green points visualize the overall progress of the predictor/corrector procedure. The black dot at the lower left is the point minimizing the constrained objective.}
\label{fig:toy_npf}
\end{figure}

\section{Small Illustrative Example}
We first illustrate our procedure with a simple problem:
\begin{align}
\begin{split}
\text{minimize} \quad x + y\label{eq:toy_npf} \\
\text{subject to} \quad x^{2} + y^{2} = 1 ,
\end{split}
\end{align}
as shown in Fig. \ref{fig:toy_npf}. For a given $(x,y)$ initialized randomly on the unit circle corresponding to the constraint $x^{2} + y^{2} = 1$, a series of predictor/corrector steps, informed by the directions specified by the objective gradient and the constraint gradient, drives the solver towards the point at which the constrained objective $x+y$ is at its minimum. The exact number of steps taken by the procedure varies and is determined by hyper-parameters such as the predictor/corrector step size. However, the overall progress of the procedure can be assessed by tracking metrics such as the angle between the objective gradient and the constraint gradient. The progress of the predictor/corrector procedure, when applied to the constrained optimization problem above, is also visualized in Fig. \ref{fig:toy_npf}. Note that the two gradients are anti-parallel at the final point.

\section{Experimental Methods}
In order to assess the generality of the method across multiple datasets and architectures, we used it on the MNIST \cite{lecun1998gradient}, CIFAR10 \cite{Krizhevsky09learningmultiple}, Iris \cite{fisher1936use}, and Auto-MPG \cite{quinlan1993combining} datasets, with both convolutional and fully connected networks. On each of these, we used the two-phase method described earlier to carry out near-zero-loss level-set traversal of the training loss function using the training split of the dataset. 


\subsection{MNIST}

The MNIST dataset corresponds to a 10-class classification problem and has 60k training examples and 10k test examples of size 28x28 pixels containing images of handwritten digits from 0 to 9. The pixel values for all examples were standardized to be in the [0, 1] interval. In order to accommodate the maximum number of experiments on large networks within the available computational capacity, all MNIST experiments used different random subsamples of 1000 examples from the full training set. The whole test set was used. 

For MNIST feed-forward (i.e., fully connected) experiments, we used a network with 3 hidden layers of 100 neurons each.  For convolutional experiments, we used two 2D convolutional layers followed by one full-connected layer (again, 3 layers total). Each convolutional layer had 20 convolutional filters of dimensions 3 $\times$ 3 and was followed by a 2D average pooling layer with pool size 2 $\times$ 2. The final feedforward layer had 10 outputs, one per class. All hidden layers used $tanh$ activation and the output layer used $softmax$. A fully-differentiable activation such as $tanh$ was a convenient choice since it guaranteed smoothness of the loss function.

In phase 1 of the predictor/corrector scheme, the weights of a network were initialized using Glorot normal initialization \cite{glorot2010understanding} and the network was trained using minibatch-gradient descent with a categorical cross-entropy loss $\mathcal{L}$ using the Adam optimizer \cite{Kingma2015AdamAM}. More specifically, the network was trained with a batch size of 32 for 100 epochs to achieve a near-zero loss on the training set of 1000 examples. 

In phase 2, the network underwent a number of predictor/corrector steps depending on whether the squared change in training loss along $\mathcal{R}$ was above or below the loss deviation threshold. $\mathcal{R}$ was set to be the squared norm function described earlier and the squared loss change threshold was set to $10^{-10}$ for all experiments for all datasets. The predictor or corrector step each had a separate full-batch optimizer, whose learning rate was adjusted heuristically based on the geometry of the landscape being traversed. More precisely, the learning rate for each optimizer was increased or decreased by a multiplicative factor depending on whether the angular change in the associated (predictor or corrector) direction in any two consecutive steps was above or below a pre-specified threshold of 0.1 degrees. The multiplicative factor was set to 0.1 and 1.1 for angular changes greater than or less than the 0.1 degrees threshold respectively. We believe the angular change heuristic takes into account the ruggedness of the loss landscapes and using it yields a less chaotic traversal of the loss landscape.  Large angular changes indicate sharp bends in the traversal curve, which call for a lower learning rate to maintain stability.

Finally, before every predictor step (after a series of corrector steps had finished), we measured training set loss, test set loss, test set accuracy, sum of squared norms for network weights, and the angle between loss gradient $\nabla_{\theta}\mathcal{L}$ and regularization gradient $\nabla_{\theta}\mathcal{R}$. In theory, phase 2 would stop when the angle between the two is $180^\circ$ or $0^\circ$, at which point there is no direction in which to decrease $\mathcal{R}$ without changing $\mathcal{L}$.  But in the numerical approximation, we observed that it was difficult to reach those exact angles in practice. For practical reasons, the number of total predictor steps taken in phase 2 was set to 10000 for MNIST feed-forward experiments and 5000 for MNIST convolutional experiments.

\subsection{CIFAR10}
The CIFAR10 dataset corresponds to a 10-class classification problem and has 60k training examples and 10k test examples which are RGB images of size 32x32. For CIFAR10 experiments too, the pixel values for all images were standardized to be in the [0, 1] interval. Similarly to the reasons described for the MNIST dataset, all CIFAR10 experiments also used different subsamples of 1000 examples from the training set. The whole test set was used.

We only report convolutional neural network experiments for CIFAR10, because traversals with fully-connected layers were less stable. This is perhaps because the learning problem underlying CIFAR10 is complex and arguably more suited for convolutional neural networks. The convolutional network used for different experiment runs had the same configuration as that for MNIST experiments: 3 hidden layers with two 2D convolutional layers, preceding a feedforward layer. However, the number of units in each hidden layer was thrice that in the MNIST case. Specifically, the convolutional layers had 60 2D convolutional filters each and the feedforward layer following them had 300 units. The dimensions for convolutional filters here were the same as those for MNIST, as were the activations and other hyper-parameters such as weight initialization method, batch-size and number of iterations for phase 1 training. The total number of predictor/corrector steps was set to 20000 for CIFAR10 experiments. 

\begin{figure}[t]
\centering
\includegraphics[width=0.8\columnwidth]{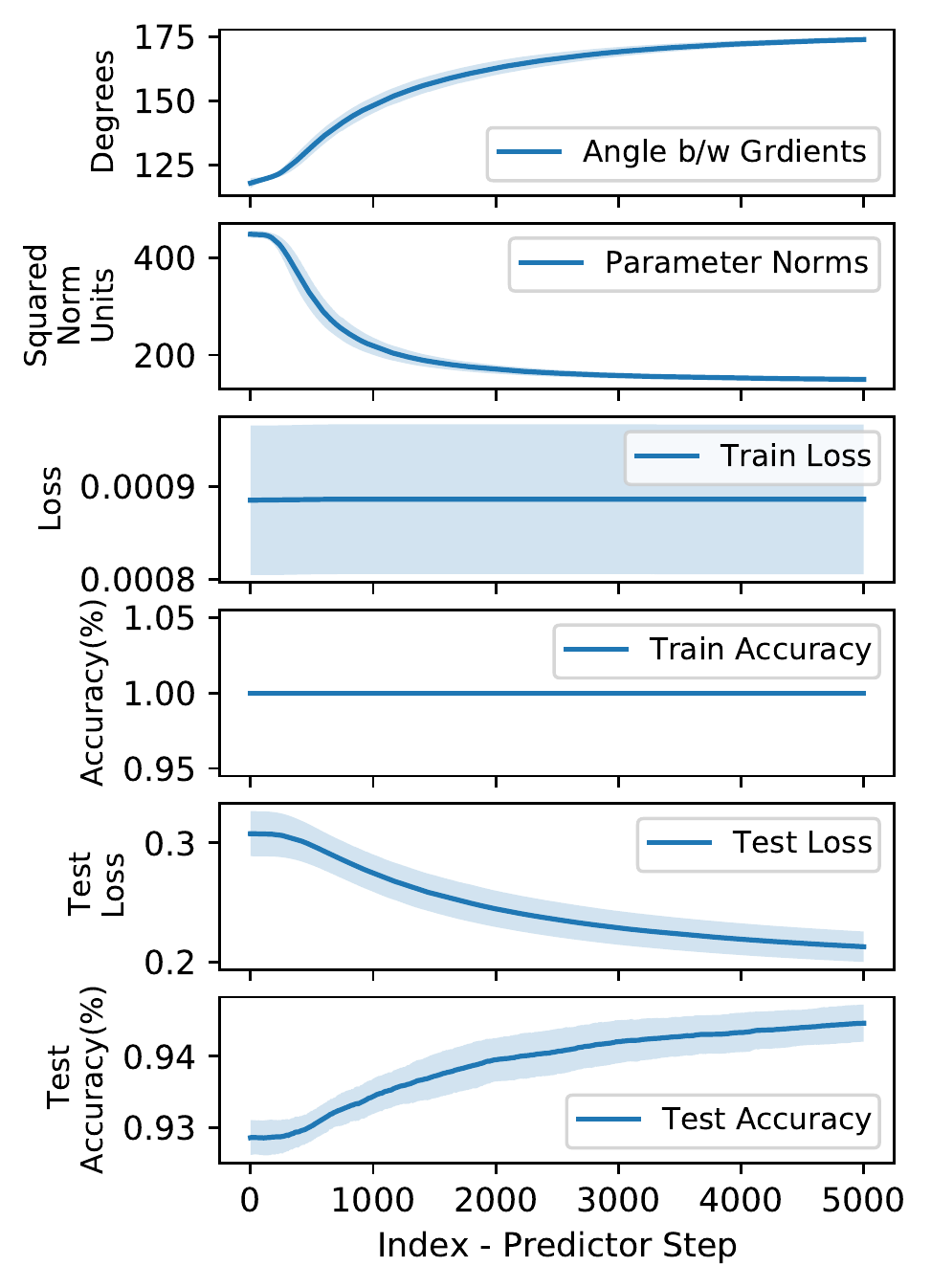}
\caption{Phase 2-learning curves for MNIST convolutional experiments.  Blue curves are averages over 10 randomly-initialized runs and the bands indicate the standard deviation.  \textbf{From top to bottom:} 1) Angle between $\nabla\mathcal{L}$ and $\nabla\mathcal{R}$ approaches $180^\circ$. 2) Squared norm of network weights decreases. 3) Training set loss remains almost constant. ) Test set loss decreases. 5) Test set accuracy increases. }
\label{fig:MNIST_CONV}
\end{figure}

\begin{figure}[t]
\centering
\includegraphics[width=0.8\columnwidth]{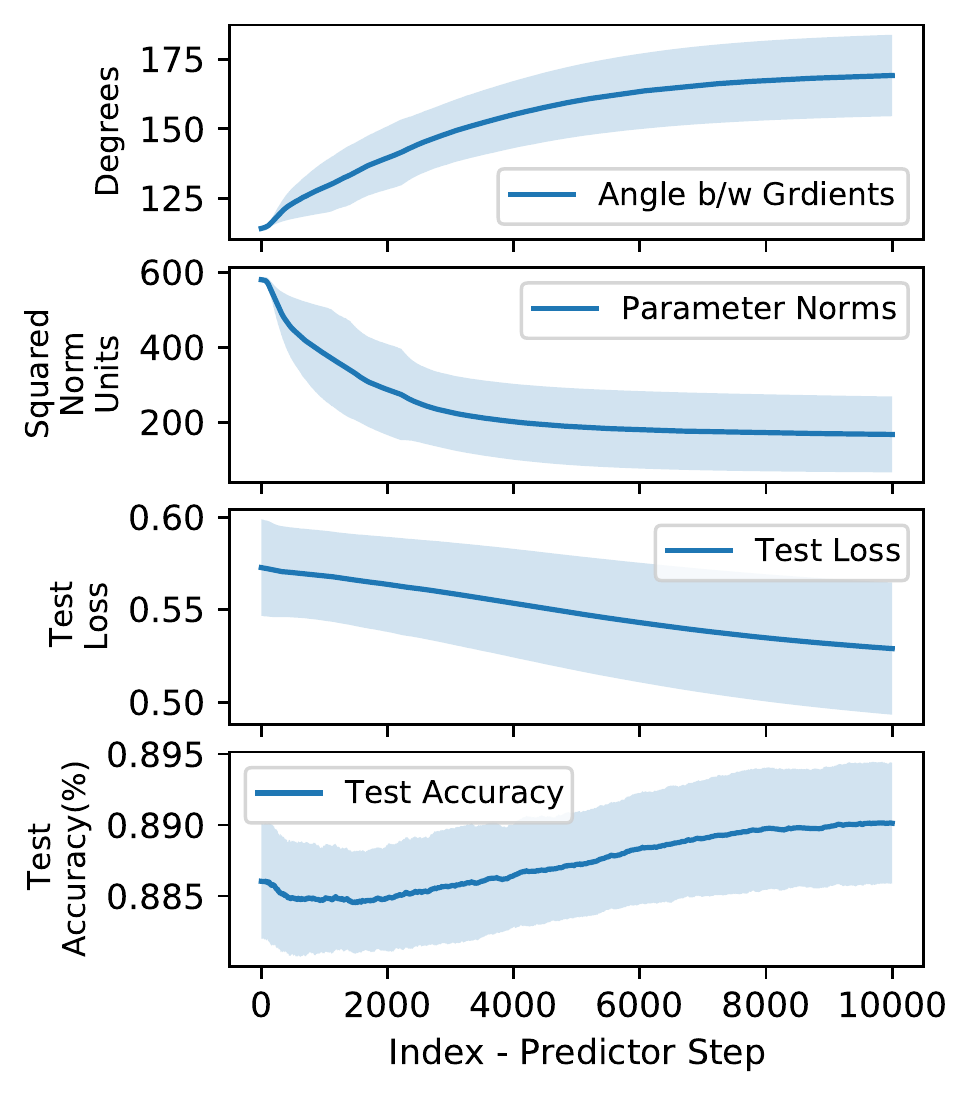}
\caption{Phase 2-learning curves for MNIST feed-forward experiments. Note that the results are aggregated over 10 random runs of the experiment.}
\label{fig:MNIST_FF}
\end{figure}

\begin{figure}[t]
\centering
\includegraphics[width=0.8\columnwidth]{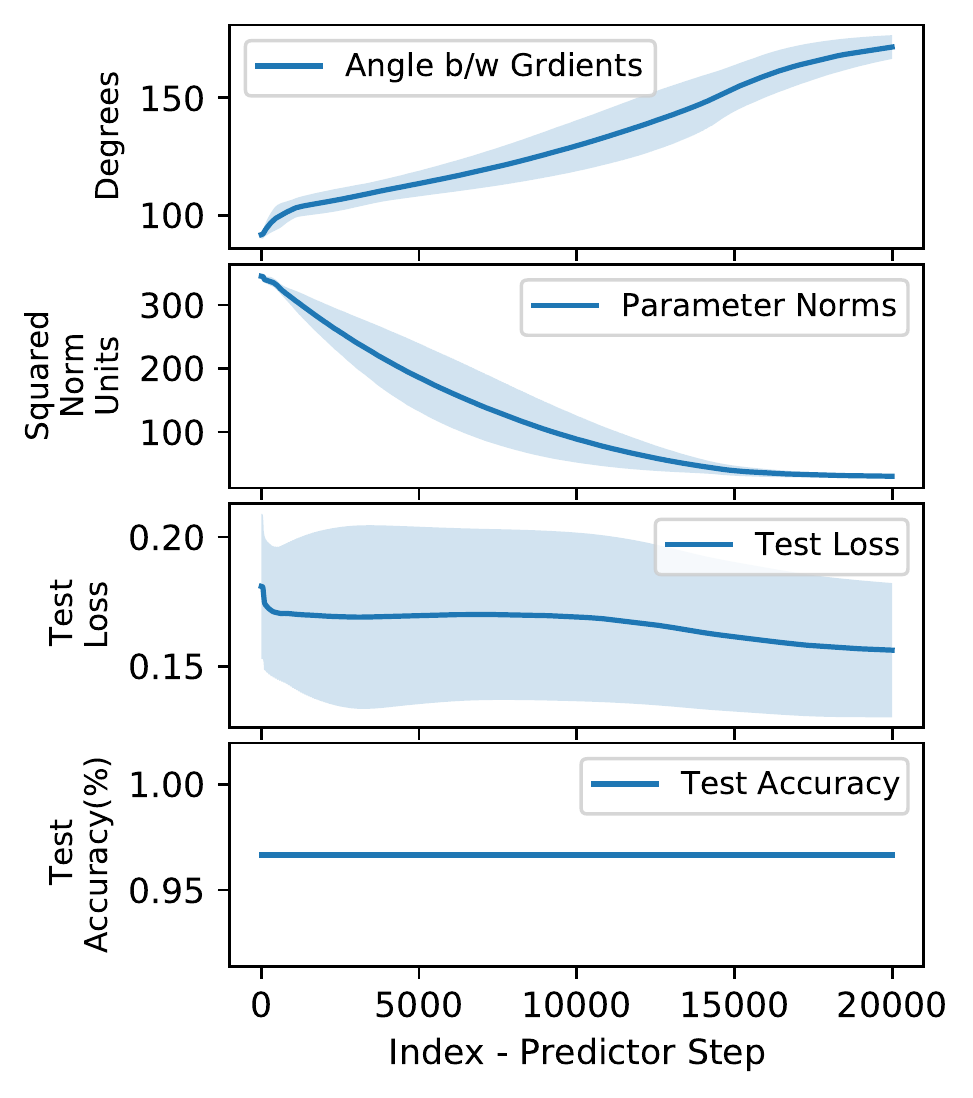}
\caption{Phase 2-learning curves for IRIS feed-forward experiments.}
\label{fig:IRIS_FF}
\end{figure}

\begin{figure}[t]
\centering
\includegraphics[width=0.8\columnwidth]{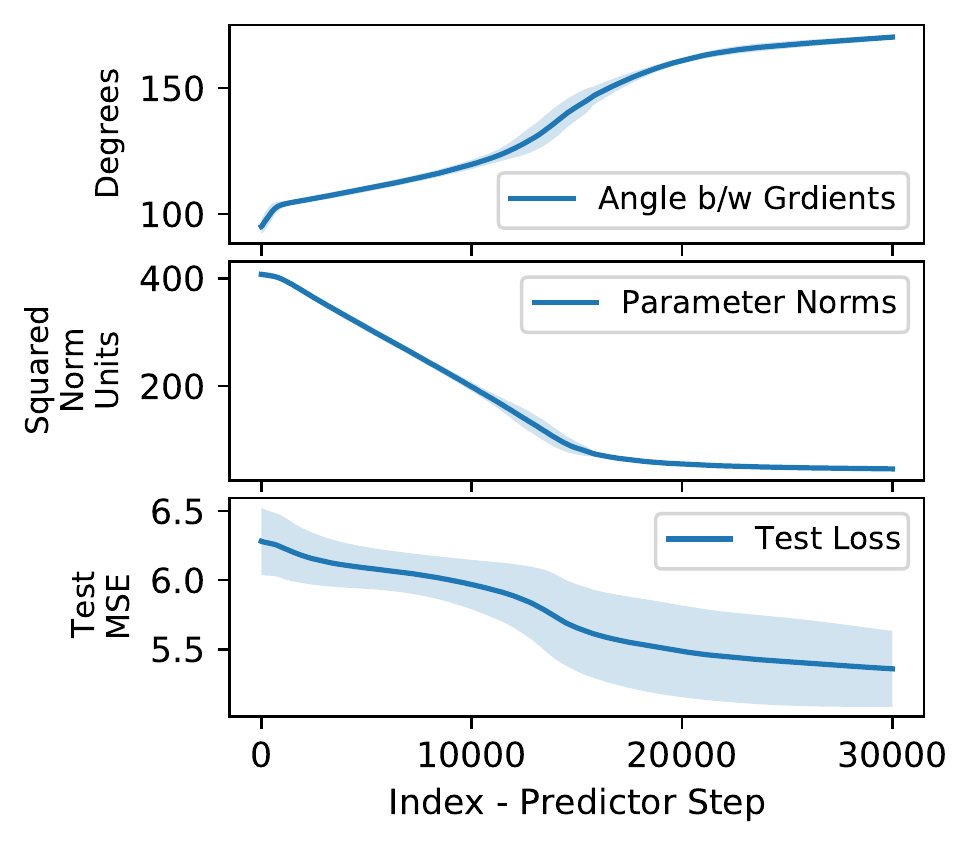}
\caption{Phase 2-learning curves for Auto-MPG feed-forward experiments.}
\label{fig:MPG_FF}
\end{figure}

\subsection{Other Datasets}
We also carried out level-set traversal experiments with Iris and Auto-MPG datasets. The Iris dataset corresponds to a 3-class classification problem and includes 150 examples. Of these examples, 120 were used as the training split and the other 30 as the test split. The Auto-MPG dataset corresponds to a regression problem and has 398 examples in total. Six of these had missing values in one of the variables and were exluded. Of the remaining 392, 314 were used as the training set and the remaining 74 were used as the test set.

For both Iris and Auto-MPG, we used a feed-forward network with 3 hidden layers with 100 neurons each. Various hyper-parameters such as weight initialization method, activations in the hidden layers etc. were the same as those in the MNIST and CIFAR10 experiments. The total number of predictor/corrector steps in phase 2 were set to 20000 and 30000 for Iris and Auto-MPG respectively. Lastly, we used mean squared error (MSE) as the loss and evaluation metric for Auto-MPG instead of average categorical cross-entropy loss and average categorical accuracy used for other datasets. This is because Auto-MPG represents a regression problem.
\begin{figure}[t]
\centering
\includegraphics[width=0.8\columnwidth]{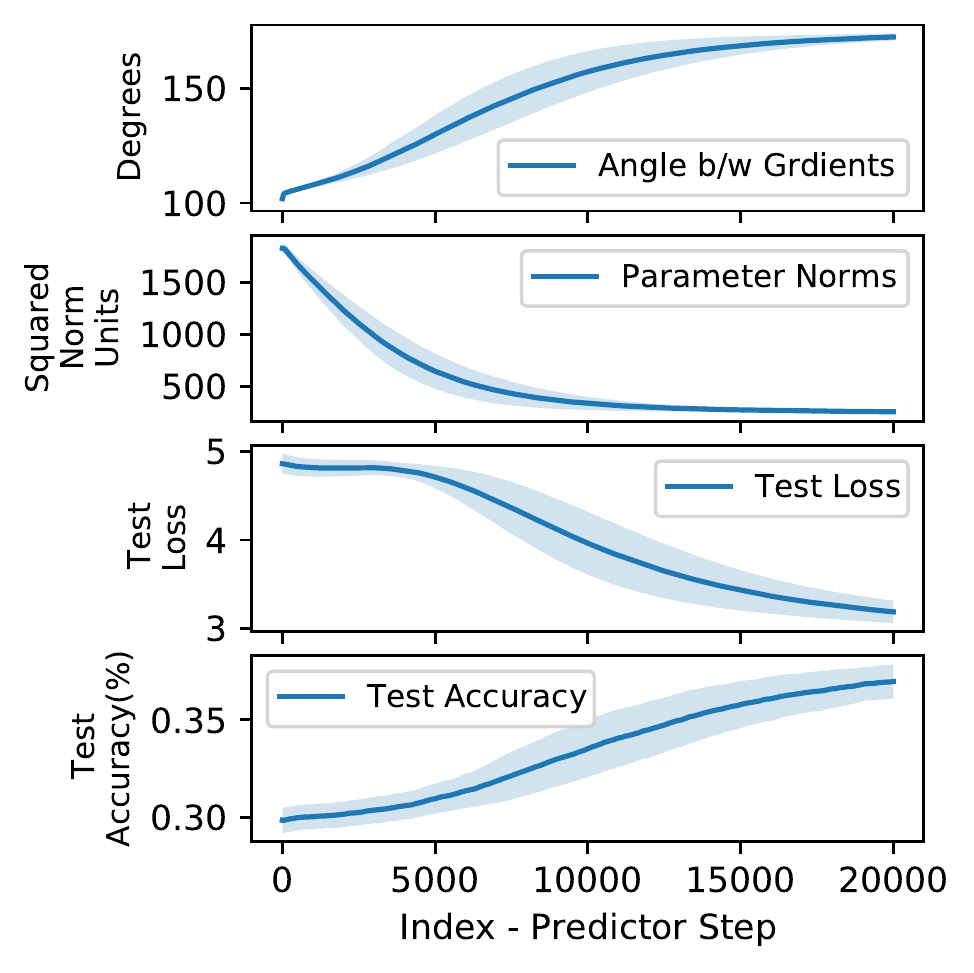}
\caption{Phase 2-learning curves for CIFAR10 convolutional experiments.}
\label{fig:CIFAR10_CONV}
\end{figure}

\subsection{Weight Decay Experiments}
We also compared the predictor/corrector approach with the more typical ``weight decay'' approach: namely, unconstrained optimization of an objective that is a weighted sum of training loss and L2-regularization. To this end, for each of the datasets described earlier, we also trained randomly-initialized instances of the networks described earlier such that the training objective now also included an L2 or squared sum of weights term scaled with a weight-decay parameter. Note that there was no two-phase training or numerical traversal in these experiments; just one phase of standard unconstrained stochastic gradient descent with an additive regularization term in the objective. The weight decay parameter, $\lambda$, is usually set through trial and error in such experiments. In our weight decay experiments, its value was taken to be in the log$_{10}$space over the interval [$10^{-6}$, $10^6$]. For each value of $\lambda$, 10 randomly-initialized networks were trained using mini-batches of 32 for an appropriate number of epochs. The total number of training epochs in these experiments was 200 for MNIST and CIFAR10. For Iris Auto-MPG datasets, the number of epochs was set to 500. The networks for MNIST and CIFAR10 were trained with random subsamples of 1000 training examples, as described earlier. 

\subsection{Results}
Since phase 1 of our approach is standard gradient-based optimization, we focus on phase 2 results. Figs. \ref{fig:MNIST_CONV} and \ref{fig:MNIST_FF} visualize the phase 2 progress over 10 random runs of MNIST convolutional and feed-forward experiments described earlier. The angle plots for these experiments show low variance over a major fraction of the traversal. This indicates relative smoothness of the landscape, which is also confirmed by the quick drop in squared norm of the network weights. Similarly, Fig. \ref{fig:CIFAR10_CONV} shows results for 10 random runs of the CIFAR10 convolutional experiment. Results for 10 random runs of Iris and Auto-MPG feed-forward experiments are shown in Figs. \ref{fig:IRIS_FF} and \ref{fig:MPG_FF} respectively. Note that different runs of the same experiment for level-set traversal used a different random subsample of the MNIST or CIFAR10 training dataset as described in the sections above. The weights of all the networks trained were also initialized randomly.

The figures show that the training loss for all new points explored during the traversal was at the same level (loss deviation being $\leq 10^{-5}$ for the specified squared deviation threshold of $10^{-10}$, to be precise) as that reached at the end of phase 1 training. This confirmed that the proposed numerical procedure almost always successfully navigates a constant near-zero-loss manifold  $\mathcal{M}$ along a regularizing $\mathcal{R}$. The traversal revealed a set of additional near-minima which displayed a generalization ability superior to or equal to that exhibited by the point in parameter space reached at the end of conventional training in phase 1. These newly-discovered near-minima almost always had lower values of the test set loss, which naturally translates to a better test set accuracy.

\begin{table}[t]
    \small
    \centering
    \begin{tabular}{ |c|c|c|c| } 
        \hline
        Experiment & Our Method & Weight-Decay & Best $\lambda$ \\
        \hline
        MNIST-CNV & $\mathbf{94.47 \pm 0.26\%}$ & 93.92 $\pm$ 0.57\% & $10^{-4}$ \\
        \hline
        MNIST-FF & 89.01 $\pm$ 0.42\% & $\mathbf{89.60 \pm 0.46\%}$ & $10^{-4}$ \\
        \hline
        CIFAR10-CNV & $\mathbf{36.96 \pm 0.87\%}$ & 35.38 $\pm$ 1.10\% & $10^{-2}$ \\
        \hline
        IRIS-FF & $\mathbf{96.67 \pm 0.0\%}$ & $\mathbf{96.67 \pm 0.0\%}$ & $10^{-3}$ \\
        \hline
        MPG-FF & $\mathbf{5.36 \pm 0.27}$ & 9.00 $\pm$ 0.73 & $10^{-4}$ \\
        \hline
    \end{tabular}
    \vspace{0.2cm}
    \caption{Comparison of Test Set Accuracy/MSE}
    \label{tab:onlytable}
\end{table}

As discussed in the previous sections, the weight decay experiments provided for a comparison between the proposed method and the one where the network loss has an added regularization term. The average test set accuracy values for 10 randomly-initialized runs of each of the MNIST feed-forward, MNIST convolutional, CIFAR10 convolutional, Iris feed-forward, and Auto-MPG feed-forward experiments at the end of level-set traversal are included in Table \ref{tab:onlytable}. Compared to the best average test set accuracy values obtained at the end of weight decay experiments runs corresponding to best-performing ($\lambda$) and which are given in the same table, the values from our method are often superior. Note that in the Auto-MPG dataset, which corresponds to a regression problem, the loss and the accuracy were both taken as mean squared error (MSE). From these results, we can argue that the proposed method on average does a better job than training on loss with added regularized term at finding well-regularized points in the loss landscape. 

 \begin{figure}[t]
\centering
\includegraphics[width=0.75\columnwidth]{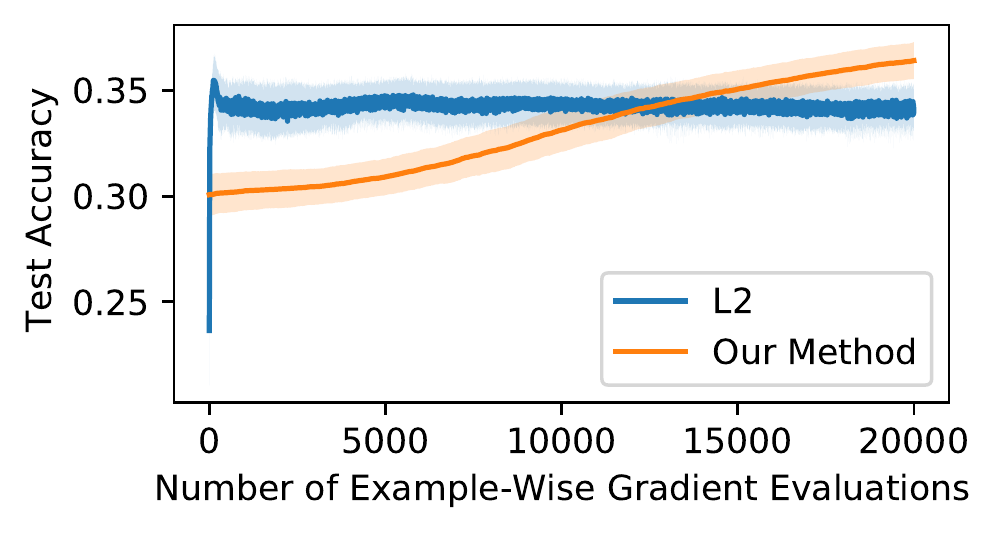}
\caption{Number of example-wise gradient computations needed to reach a given performance level on CIFAR10, for weight decay (L2) and traversal (our method).}
\label{fig:complex}
\end{figure}

\section{Computational Complexity}
The main computational expense in both our method and standard weight decay comes from evaluation of the loss and its gradient on many training examples.  Therefore, their relative computational complexities depend on the number of example-wise gradient evaluations needed to reach a given level of performance.  Empirically, this number is highly problem-specific.  On Iris, MPG, and MNIST, our method was favorable regardless of the number of gradient evaluations.  In other words, for any number of additional gradient evaluations after phase 1, our method is always above, or within one standard deviation of, weight decay testing performance.  The CIFAR10 benchmark was more interesting, as shown in Fig.\@ \ref{fig:complex}.  In this case, weight decay initially outperforms our method substantially, but does not improve with additional gradient evaluations.  In contrast, our method steadily improves with additional computation, and outperforms weight decay after $\sim$12.5K example-wise gradient evaluations in phase 2.

 \begin{figure}[t]
\centering
\includegraphics[width=0.8\columnwidth]{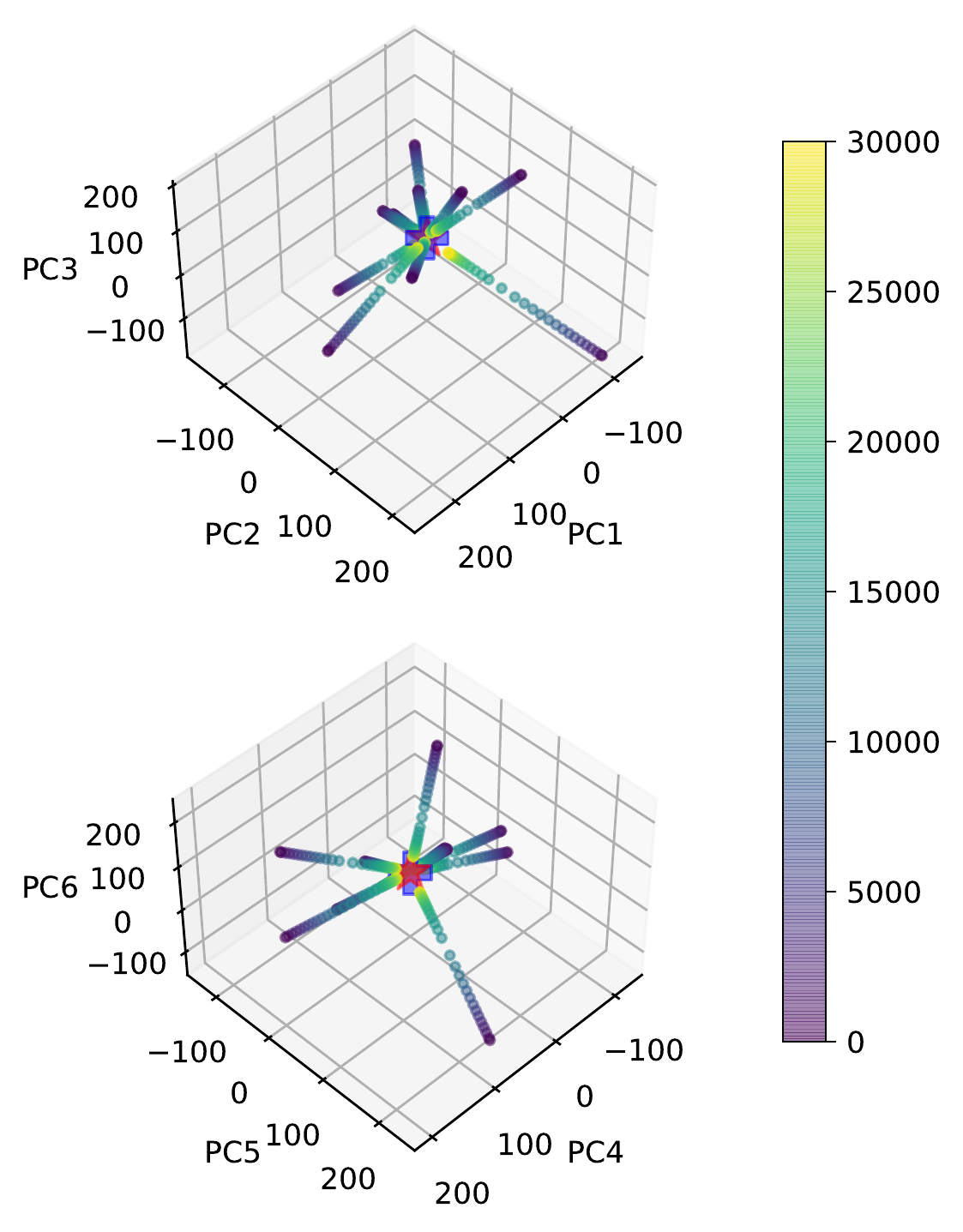}
\caption{Visualizing 10 randomly-initialized level-set traversals for MPG feed-forward experiments. The traversed points have been color-coded from first to last as visualized in the appended colorbar. The mean of endpoints for the traversals is given as blue cross while the mean of points obtained at the end of training with weight decay for best-performing $\lambda$ is given as red star. Note that the latter two points are almost collocated (in the top-6 principal subspace).}
\label{fig:total_traversal}
\end{figure}

\section{Visualizing Level-Set Traversals}

We also used dimensionality reduction to provide visual insight into the level-set trajectories. More specifically, we used Principal Component Analysis (PCA) \cite{pearson1901liii} to obtain projections of these trajectories along the top 6 principal components with the most explained variance ratio. PCA was carried out on the set of weight vectors obtained by subsampling every 5$^{th}$ point of a phase 2 traversal. The trajectories, when plotted along with the points obtained at the end of training with weight decay, provided for a visual confirmation of both methods converging to the same highly-regularized neighborhood in the parameter space.

Fig. \ref{fig:total_traversal} visualizes the whole level-set trajectories for 10 random runs of MNIST feed-forward experiments along with the arithmetic mean of the weight vectors obtained at the end of 10 runs of weight decay experiments corresponding to the best-performing $\lambda$ (red star). The average of the end-points of the trajectories (blue cross), which occurs very close to the average end-point for training with weight decay, is also shown. Note that the top 6 principle components are split into two 3D sub-plots. As seen in the figure, all traversals seem to be moving to the same neighborhood. This indicates the presence of a global heavily-regularized neighborhood in the parameter-space. Based on intuition from stochastic weight averaging \cite{Izmailov2018AveragingWL}, we hypothesized that better test performance might be achieved by collecting the final points from each traversal, and then averaging them to produce a single new, superior weight vector (blue cross).  In fact, the measured loss at this average weight vector was worse, not better, than at the final traversal points.  This suggests some non-convexity in the loss surface near the highly regularized points.
 
\begin{figure}[h!]
\centering
\includegraphics[width=0.7\columnwidth]{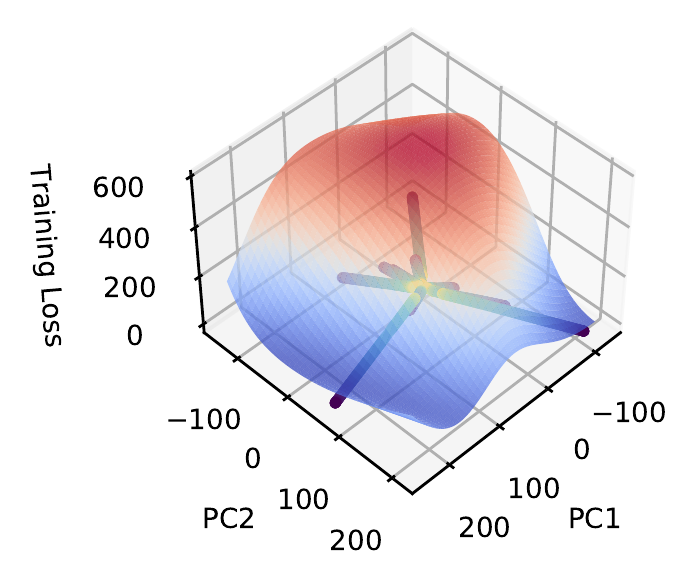}
\caption{Visualizing network training loss for the MPG dataset over a 100 $\times$ 100 grid sampled in the top-2 PCA sub-space. Level-set trajectories for randomly-initialized networks approach an area with substantially higher training loss. The fraction of total variance explained by the top two principal components is 21.04\%.}
\label{fig:total_traversal_loss_mpg}
\end{figure}

To test this hypothesis, we visualized auto-MPG training loss surface in the top-2 principle component sub-space in Fig.\@ \ref{fig:total_traversal_loss_mpg}.  Regularly-spaced grid points in this sub-space were inverse-transformed to weight space to evaluate training loss, and trajectory points were also projected into this sub-space.
 
\begin{figure}[h!]
\centering
\includegraphics[width=0.7\columnwidth]{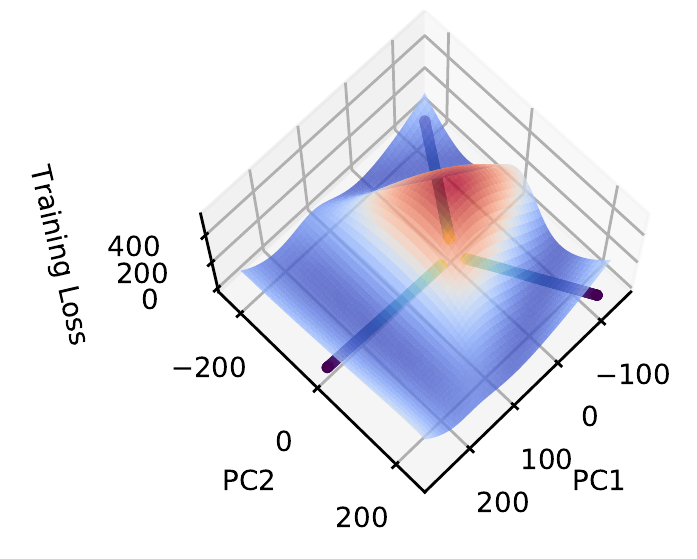}
\caption{Like Fig.\@ \ref{fig:total_traversal_loss_mpg}, but with only three traversals. The fraction of total variance explained by the top two principal components is 79.93\%. }
\label{fig:subset_traversal_loss_mpg}
\end{figure}

  \begin{figure}[h!]
\centering
\includegraphics[width=0.7\columnwidth]{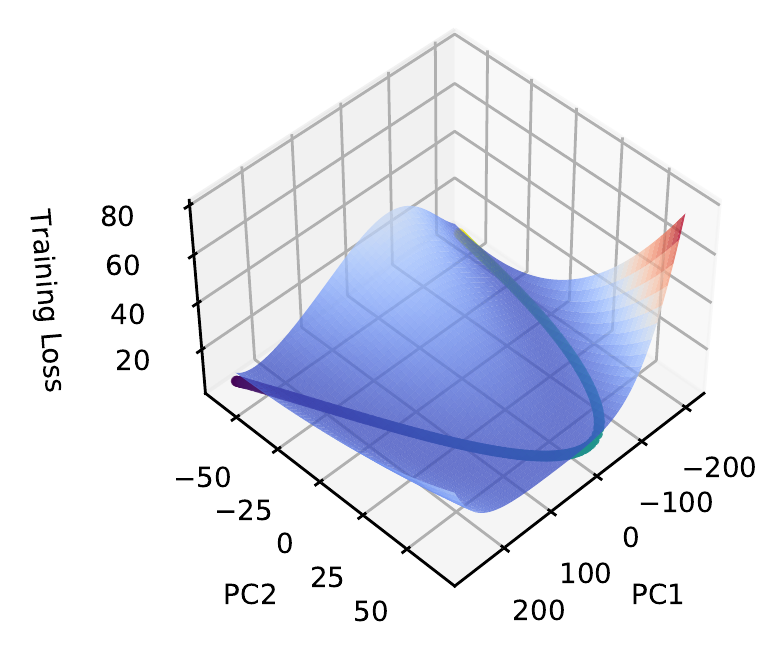}
\caption{Like Fig.\@ \ref{fig:total_traversal_loss_mpg}, but with only one traversal. The fraction of total variance explained by the top two principal components is 97.01\%. }
\label{fig:single_traversal_loss_mpg}
\end{figure}

The loss surface in Fig.\@ \ref{fig:total_traversal_loss_mpg} does not perfectly reflect the fixed $\epsilon$ loss of the level-set traversal trajectories, because the inverse-transformed grid points do not perfectly coincide with the traversals, which have unexplained variance outside the top-2 PCA sub-space.  However, limiting PCA to fewer than 10 traversals can reduce this unexplained variance and produce more accurate loss surface visualizations.  Using only 3 or 1 traversals (Figs.\@  \ref{fig:subset_traversal_loss_mpg} and \ref{fig:single_traversal_loss_mpg}, respectively), we see the loss surface visualization is closer to zero near the traversal points.


\section{Discussion}
We have presented a novel two-phased numerical method for locating almost optimally regularized weights within the manifold of near-zero training loss in deep neural networks. We applied the method on various multivariate classification and regression benchmarks and architectures. When applied to the specific problem of minimizing a regularizer at a constant loss level, our method performs better on average than the more typical weight decay approach. Furthermore, we use dimensionality-reduction methods to visualize the loss dynamics over these level-set traversals.

Since the method uses only first-order information about the loss and constraint functions, it can arguably be used on medium to large datasets and networks. Even though we worked with subsets of the MNIST and CIFAR10 datasets in order to conduct maximum number of experiments within available compute, the calculations of full-batch update, as utilized in the predictor/corrector steps, can be readily distributed over multiple GPUs, using standard provisions in modern deep learning libraries such as Google's TensorFlow \cite{DBLP:conf/mlsys/AgrawalMPLASGLH19,nilsen2019efficient}.

While we have reported results for numerical exploration of constant, near-zero loss manifolds along a regularizing path only, the method presented here can possibly be generalized to explore constant-loss manifolds along differentiable paths of interest in deep neural networks. This is akin to multi-objective optimization where optimizing successive objectives improves the overall quality of solutions found. For our future works, we aim to extend the method to such multi-objective problems in the context of machine learning and deep neural networks. We also seek to investigate the scenarios where loss is `exactly' zero, rendering first-order methods insufficient. While second-order methods are expensive and borderline impractical for large networks, some recent works exploring the possible existence of a structure in the Hessian of large networks \cite{sagun2018empirical} provide for an interesting starting point in this regard.

\bibliographystyle{IEEEtran}
\bibliography{IEEEabrv,mybib}

\begin{thebibliography}{10}
\providecommand{\url}[1]{#1}
\csname url@samestyle\endcsname
\providecommand{\newblock}{\relax}
\providecommand{\bibinfo}[2]{#2}
\providecommand{\BIBentrySTDinterwordspacing}{\spaceskip=0pt\relax}
\providecommand{\BIBentryALTinterwordstretchfactor}{4}
\providecommand{\BIBentryALTinterwordspacing}{\spaceskip=\fontdimen2\font plus
\BIBentryALTinterwordstretchfactor\fontdimen3\font minus
  \fontdimen4\font\relax}
\providecommand{\BIBforeignlanguage}[2]{{%
\expandafter\ifx\csname l@#1\endcsname\relax
\typeout{** WARNING: IEEEtran.bst: No hyphenation pattern has been}%
\typeout{** loaded for the language `#1'. Using the pattern for}%
\typeout{** the default language instead.}%
\else
\language=\csname l@#1\endcsname
\fi
#2}}
\providecommand{\BIBdecl}{\relax}
\BIBdecl

\bibitem{poole2016exponential}
B.~Poole, S.~Lahiri, M.~Raghu, J.~Sohl-Dickstein, and S.~Ganguli, ``Exponential
  expressivity in deep neural networks through transient chaos,'' in
  \emph{Advances in Neural Information Processing Systems}, 2016, pp.
  3360--3368.

\bibitem{zagoruyko2016wide}
S.~Zagoruyko and N.~Komodakis, ``Wide residual networks,'' in \emph{British
  Machine Vision Conference 2016}.\hskip 1em plus 0.5em minus 0.4em\relax
  British Machine Vision Association, 2016.

\bibitem{DBLP:journals/corr/SoudryC16}
\BIBentryALTinterwordspacing
D.~Soudry and Y.~Carmon, ``No bad local minima: Data independent training error
  guarantees for multilayer neural networks,'' \emph{CoRR}, vol.
  abs/1605.08361, 2016. [Online]. Available:
  \url{http://arxiv.org/abs/1605.08361}
\BIBentrySTDinterwordspacing

\bibitem{baldi2019capacity}
P.~Baldi and R.~Vershynin, ``The capacity of feedforward neural networks,''
  \emph{Neural networks}, vol. 116, pp. 288--311, 2019.

\bibitem{du2019gradient}
S.~Du, J.~Lee, H.~Li, L.~Wang, and X.~Zhai, ``Gradient descent finds global
  minima of deep neural networks,'' in \emph{International Conference on
  Machine Learning}, 2019, pp. 1675--1685.

\bibitem{jaeger2004harnessing}
H.~Jaeger and H.~Haas, ``Harnessing nonlinearity: Predicting chaotic systems
  and saving energy in wireless communication,'' \emph{Science}, vol. 304, no.
  5667, pp. 78--80, 2004.

\bibitem{huang2006extreme}
G.-B. Huang, Q.-Y. Zhu, and C.-K. Siew, ``Extreme learning machine: theory and
  applications,'' \emph{Neurocomputing}, vol.~70, no. 1-3, pp. 489--501, 2006.

\bibitem{eliasmith2004neural}
C.~Eliasmith and C.~H. Anderson, \emph{Neural engineering: Computation,
  representation, and dynamics in neurobiological systems}.\hskip 1em plus
  0.5em minus 0.4em\relax MIT press, 2004.

\bibitem{rosen1960gradient}
J.~B. Rosen, ``The gradient projection method for nonlinear programming. part
  i. linear constraints,'' \emph{Journal of the Society for Idustrial and
  Applied Mathematics}, vol.~8, no.~1, pp. 181--217, 1960.

\bibitem{allgower1997numerical}
E.~L. Allgower and K.~Georg, ``Numerical path following,'' \emph{Handbook of
  Numerical Analysis}, vol.~5, no.~3, p. 207, 1997.

\bibitem{lecun1998gradient}
Y.~LeCun, L.~Bottou, Y.~Bengio, P.~Haffner \emph{et~al.}, ``Gradient-based
  learning applied to document recognition,'' \emph{Proceedings of the IEEE},
  vol.~86, no.~11, pp. 2278--2324, 1998.

\bibitem{Krizhevsky09learningmultiple}
A.~Krizhevsky, ``Learning multiple layers of features from tiny images,'' Tech.
  Rep., 2009.

\bibitem{fisher1936use}
R.~A. Fisher, ``The use of multiple measurements in taxonomic problems,''
  \emph{Annals of Eugenics}, vol.~7, no.~2, pp. 179--188, 1936.

\bibitem{quinlan1993combining}
J.~R. Quinlan, ``Combining instance-based and model-based learning,'' in
  \emph{Proceedings of the Tenth International Conference on Machine Learning},
  1993, pp. 236--243.

\bibitem{sagun2018empirical}
L.~Sagun, U.~Evci, V.~U. Guney, Y.~Dauphin, and L.~Bottou, ``Empirical analysis
  of the hessian of over-parametrized neural networks,'' in \emph{International
  Conference on Learning Representations}, 2018.

\bibitem{venturi2019spurious}
L.~Venturi, A.~S. Bandeira, and J.~Bruna, ``Spurious valleys in
  one-hidden-layer neural network optimization landscapes,'' \emph{The Journal
  of Machine Learning Research}, vol.~20, no. 133, pp. 1--34, 2019.

\bibitem{DBLP:conf/icml/Nguyen19}
Q.~Nguyen, ``On connected sublevel sets in deep learning,'' in
  \emph{Proceedings of the 36th International Conference on Machine Learning,
  {ICML} 2019, 9-15 June 2019, Long Beach, California, {USA}}, ser. Proceedings
  of Machine Learning Research, K.~Chaudhuri and R.~Salakhutdinov, Eds.,
  vol.~97.\hskip 1em plus 0.5em minus 0.4em\relax {PMLR}, 2019, pp. 4790--4799.

\bibitem{DBLP:conf/icml/DraxlerVSH18}
\BIBentryALTinterwordspacing
F.~Draxler, K.~Veschgini, M.~Salmhofer, and F.~A. Hamprecht, ``Essentially no
  barriers in neural network energy landscape,'' in \emph{Proceedings of the
  35th International Conference on Machine Learning, {ICML} 2018,
  Stockholmsm{\"{a}}ssan, Stockholm, Sweden, July 10-15, 2018}, ser.
  Proceedings of Machine Learning Research, J.~G. Dy and A.~Krause, Eds.,
  vol.~80.\hskip 1em plus 0.5em minus 0.4em\relax {PMLR}, 2018, pp. 1308--1317.
  [Online]. Available: \url{http://proceedings.mlr.press/v80/draxler18a.html}
\BIBentrySTDinterwordspacing

\bibitem{liu2017learning}
Z.~Liu, J.~Li, Z.~Shen, G.~Huang, S.~Yan, and C.~Zhang, ``Learning efficient
  convolutional networks through network slimming,'' in \emph{Proceedings of
  the IEEE International Conference on Computer Vision}, 2017, pp. 2736--2744.

\bibitem{lecun1990optimal}
Y.~LeCun, J.~S. Denker, and S.~A. Solla, ``Optimal brain damage,'' in
  \emph{Advances in Neural Information Processing Systems}, 1990, pp. 598--605.

\bibitem{srivastava2014dropout}
N.~Srivastava, G.~Hinton, A.~Krizhevsky, I.~Sutskever, and R.~Salakhutdinov,
  ``Dropout: A simple way to prevent neural networks from overfitting,''
  \emph{The Journal of Machine Learning Research}, vol.~15, no.~1, pp.
  1929--1958, 2014.

\bibitem{garipov2018loss}
T.~Garipov, P.~Izmailov, D.~Podoprikhin, D.~P. Vetrov, and A.~G. Wilson, ``Loss
  surfaces, mode connectivity, and fast ensembling of dnns,'' in \emph{Advances
  in Neural Information Processing Systems}, 2018, pp. 8789--8798.

\bibitem{jonsson1998nudged}
H.~Jonsson, ``Nudged elastic band method for finding minimum energy paths of
  transitions,'' \emph{Classical and Quantum Dynamics in Condensed Phase
  Simulations}, vol. 385, 1998.

\bibitem{boyd2004convex}
S.~Boyd, S.~P. Boyd, and L.~Vandenberghe, \emph{Convex optimization}.\hskip 1em
  plus 0.5em minus 0.4em\relax Cambridge university press, 2004.

\bibitem{glorot2010understanding}
X.~Glorot and Y.~Bengio, ``Understanding the difficulty of training deep
  feedforward neural networks,'' in \emph{Proceedings of the Thirteenth
  International Conference on Artificial Intelligence and Statistics}, 2010,
  pp. 249--256.

\bibitem{Kingma2015AdamAM}
D.~P. Kingma and J.~Ba, ``Adam: A method for stochastic optimization,''
  \emph{CoRR}, vol. abs/1412.6980, 2015.

\bibitem{pearson1901liii}
K.~Pearson, ``Liii. on lines and planes of closest fit to systems of points in
  space,'' \emph{The London, Edinburgh, and Dublin Philosophical Magazine and
  Journal of Science}, vol.~2, no.~11, pp. 559--572, 1901.

\bibitem{Izmailov2018AveragingWL}
P.~Izmailov, D.~Podoprikhin, T.~Garipov, D.~Vetrov, and A.~Wilson, ``Averaging
  weights leads to wider optima and better generalization,'' in
  \emph{Conference on Uncertainty in Artificial Intelligence (UAI)}, 2018.

\bibitem{DBLP:conf/mlsys/AgrawalMPLASGLH19}
A.~Agrawal, A.~N. Modi, A.~Passos, A.~Lavoie, A.~Agarwal, A.~Shankar,
  I.~Ganichev, J.~Levenberg, M.~Hong, R.~Monga, and S.~Cai, ``Tensorflow eager:
  {A} multi-stage, python-embedded {DSL} for machine learning,'' in
  \emph{Proceedings of Machine Learning and Systems 2019}, A.~Talwalkar,
  V.~Smith, and M.~Zaharia, Eds., 2019.

\bibitem{nilsen2019efficient}
G.~K. Nilsen, A.~Z. Munthe-Kaas, H.~J. Skaug, and M.~Brun, ``Efficient
  computation of hessian matrices in tensorflow,'' \emph{arXiv preprint
  arXiv:1905.05559}, 2019.

\end{thebibliography}

\end{document}